\pdfoutput=1
\documentclass[11pt,preprint]{article}
\usepackage{adjustbox}
\usepackage{acl}
\usepackage{booktabs} 
\usepackage{times}
\usepackage[utf8]{inputenc}
\usepackage[T1]{fontenc}
\usepackage{graphicx}
\usepackage{subcaption}
\usepackage{amsmath}

\usepackage{times}
\usepackage{latexsym}

\usepackage[T1]{fontenc}

\usepackage[utf8]{inputenc}

\usepackage{microtype}

\usepackage{inconsolata}

\usepackage{graphicx}

\title{Probing the phon* representations of the Mimi codebooks}

\author{\rm
Artem Saloev$^1$,
Nicolas Ballier$^1$,
Erin Pacquetet$^2$\\
$^1$ALTAE, Universit{\'e} Paris Cit{\'e}, F-75013 Paris, France\\
$^2$SCIAM, 10 rue de Penthièvre, F-75008 Paris, France\\
    \texttt{artem.saloev@gmail.com, nicolas.ballier@u-pariscite.fr,}\\ 
    \texttt{erin.pacquetet@sciam.fr}
}

\begin{document}
\maketitle
\begin{abstract}
Neural audio codecs are increasingly used to encode speech for
integration with large language models. This paper investigates the
phonetic representations in the first codebook (CB0) of the Mimi
codec. ABX-style evaluation is poorly suited to Mimi's fixed 80\,ms
tokenization window, which often spans multiple phonemes; we
instead use forced alignment, entropy analysis, and Phone-Normalized
Mutual Information (PNMI). Using TIMIT speech (58/55 speakers,
SA1/SA2), individual tokens are strongly associated with specific
phonemes (PNMI~$=0.84$, $3.8\times$ above random), yet phonemes
are realized by multiple competing tokens rather than a one-to-one
mapping. A permutation test confirms token distributions differ
significantly across sentences (JSD~$=0.61$ vs.\ null $=0.14$).
PNMI analysis across all eight codebooks shows phonological
information is distributed throughout the RVQ hierarchy, suggesting
CB0 implements a context sensitive allophonic encoding rather than
a stable phoneme inventory.
\end{abstract}

\section{Introduction}

Neural audio codecs (NACs) have emerged as a fundamental component of recent spoken language systems, providing a discrete interface between continuous speech signals and large language models. Architectures such as SoundStream \citep{zeghidour2021soundstream}, EnCodec \citep{defossez2022high}, and Mimi \citep{defossez2024moshi} compress speech into sequences of discrete tokens while preserving perceptual quality. These representations are increasingly employed not only for compression, but also as inputs to speech generation and conversational systems. Despite their widespread use, the linguistic content encoded by these discrete units remains poorly understood.

\begin{figure}
    \centering
    \includegraphics[scale=0.3]{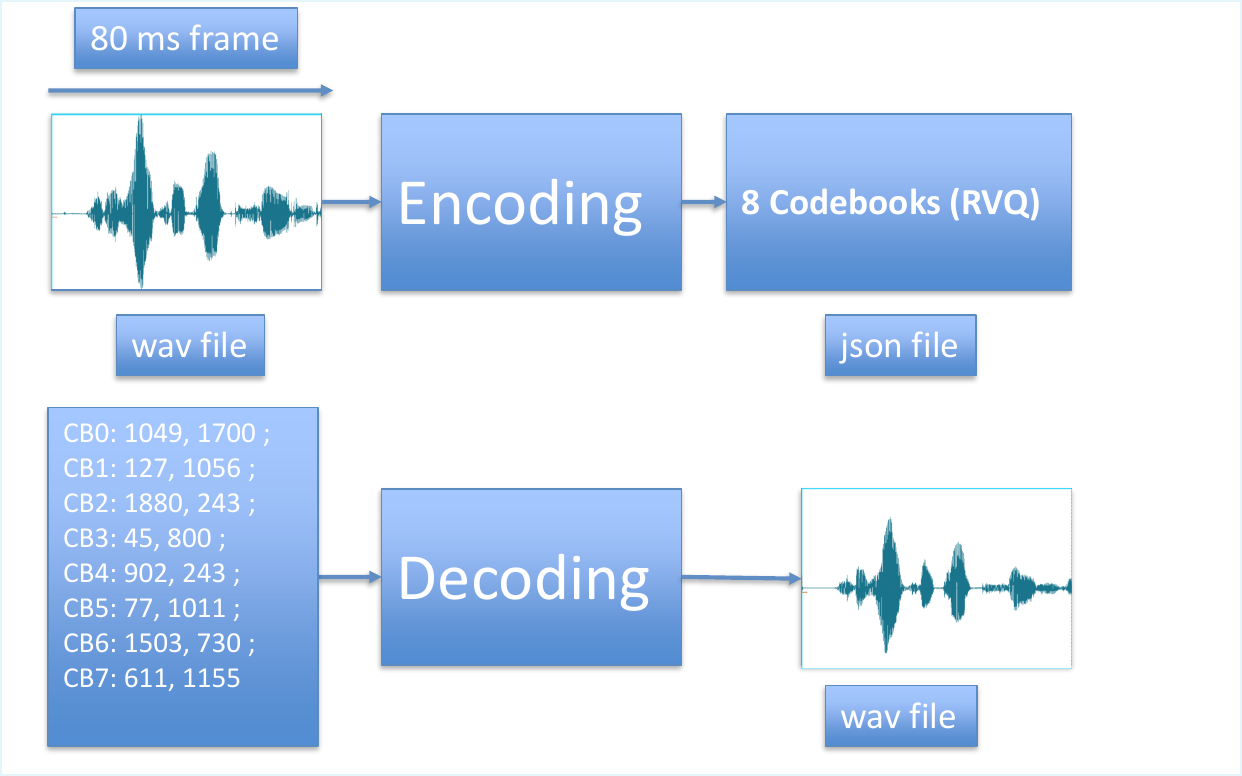}
    \caption{A schematic representation of the Mimi encoding-decoding pipeline}
    \label{fig:Mimipipeline}
\end{figure}

Recent surveys have highlighted the unresolved question of whether codec tokens correspond to interpretable linguistic units such as phonemes, subphonemic realizations, prosodic patterns, or broader semantic structures \citep{cui2025recent,mousavi2025discrete,ye2025codec}. Existing evaluations have frequently relied on ABX-style discrimination tasks inherited from self-supervised speech representation learning. However, these methods implicitly assume that the underlying representations can be localized to short segment-sized events. This assumption is difficult to reconcile with neural codecs such as Mimi, whose tokens correspond to fixed 80ms frames of speech and can therefore span multiple phonetic events. Rather than treating codec tokens as isolated segmental representations, we adopt a distributional perspective that examines their statistical relationship with phonetic categories while explicitly accounting for token--phoneme overlap. Mimi employs residual vector quantization with eight codebooks operating at a frame rate of 12.5 frames per second \citep{defossez2024moshi}. Figure~\ref{fig:Mimipipeline} provides a simplified overview of the Mimi encoding and decoding process. The Mimi developers suggest that the first codebook captures relatively high-level information related to semantic content, while the remaining codebooks contribute complementary information required for faithful speech reconstruction. However, the precise representational role of this initial codebook remains unclear. Does it behave as a stable inventory of phoneme-like units, or does it instead preserve finer contextual variation within phonemic categories? This question motivates the analyses presented in this work.

To address this question, we analyze the first Mimi codebook using an entropy-based probing framework inspired by posteriorgram analyses \citep{churchwell2024high} and by the bidirectional token--phoneme methodology introduced for SpeechTokenizer \citep{zhang2024speechtokenizer}. We characterize the asymmetry of token--phoneme associations, investigate the stability of a token family across sentences using a phoneme-conditioned permutation test, and evaluate the effect of manipulating the first codebook through controlled interventions. Using speech from the TIMIT corpus \citep{garofolo1993darpa}, we show that although individual tokens exhibit strong preferences for specific phonetic categories, phonemes themselves are realized through distributed and context-sensitive token families. These findings suggest that the first Mimi codebook is not a simple phoneme inventory but instead captures richer acoustic structure relevant to speech reconstruction.

\section{Previous Research}

\subsection{Neural Audio Codecs}

Neural audio codecs (NACs) transform continuous speech signals into sequences of discrete tokens, providing an efficient interface between speech and transformer-based architectures. SoundStream \citep{zeghidour2021soundstream} introduced residual vector quantization (RVQ), in which successive quantizers encode the residual information left by preceding stages, enabling low-bitrate compression while preserving perceptual quality. EnCodec \citep{defossez2022high} further improved this paradigm through adversarial training and multi-band processing, demonstrating that discrete audio representations can support high-quality speech reconstruction.

More recently, NACs have evolved from compression systems into \emph{speech tokenizers} for spoken language models \citep{Borsos2023SoundStorm,Jiang2024UniversalTokens,li2025dualcodec}. In these systems, codec tokens serve as the intermediate representation through which speech is processed by downstream models. Consequently, the properties of these tokens directly influence which aspects of the speech signal are preserved and made accessible to subsequent stages of processing. Despite their increasing importance, the linguistic interpretation of codec tokens remains poorly understood. Do they correspond to phonemes, finer-grained subphonemic realizations, prosodic events, or combinations of multiple representational levels? Addressing this question is essential for understanding the capabilities and limitations of contemporary spoken language systems. 

\subsection{Previous Research on Mimi}

Mimi is a recent neural audio codec based on residual vector quantization (RVQ) that produces eight codebooks at a frame rate of 12.5,Hz, corresponding to fixed 80ms segments of speech \citep{defossez2024moshi}. The architecture achieves high compression while preserving the perceptual quality of reconstructed speech, making it suitable for contemporary spoken language systems. In the original description of Mimi, the first codebook (\texttt{codebook\_0}) is presented as carrying the most semantically relevant information through distillation from the self-supervised speech model WavLM \citep{chen2022wavlm}. In this context, ``semantic'' does not refer strictly to lexical meaning, but rather to representations expected to remain more stable across acoustic variation and to facilitate language-model prediction over token sequences. The remaining codebooks contribute complementary acoustic information required for accurate waveform reconstruction.

Despite the growing adoption of Mimi, the representational status of its discrete units remains largely unexplored. In particular, it is unclear whether tokens from the first codebook correspond to stable phoneme-like categories, finer-grained subphonemic realizations, or broader context-dependent patterns of speech organization. Recent discussions of codec-based spoken language models have highlighted the importance of the underlying discrete representation for downstream performance \citep{ye2025codec}, yet they have not provided a detailed phonetic analysis of Mimi's token sequences.  \citep{ballier26_speechprosody} argued that Mimi's last seven codebooks partially encoded prosodic features such as pitch, but, to our knowledge, no previous work has systematically investigated the token--phoneme organization of Mimi's first codebook using distributional probing and controlled interventions. This gap motivates the present study. 

\subsection{The SpeechTokenizer Method}
SpeechTokenizer is a neural speech tokenizer based on residual vector quantization that learns hierarchical discrete representations of speech \citep{zhang2024speechtokenizer}. Through a series of analyses, Zhang et al. investigated the linguistic content of the learned codebooks and reported that the first quantization layer captured a substantial portion of the phonetic information present in the signal.

Of particular relevance to the present work is the Codebook Analysis procedure introduced in Appendix F of SpeechTokenizer. This analysis combines forced phonetic alignments with token sequences to estimate statistical associations between discrete units and phoneme categories. By examining conditional token--phoneme distributions derived from token–phoneme co-occurrence statistics, the method provides a framework for evaluating whether discrete codec representations exhibit phonetic organization. SpeechTokenizer further visualizes these associations using probability heatmaps and reports Phone-Normalized Mutual Information (PNMI) as a global measure of correspondence between tokens and phonemes. In the present study, we adopt the distributional perspective introduced by this analysis while extending it to investigate entropy asymmetries, token-family structure, cross-sentence stability, and controlled interventions within Mimi's first codebook.

\subsection{Limitations of ABX Evaluation for Fixed-Frame Codecs}

ABX discrimination tasks assess whether a representation system can
distinguish between two phonetic categories $A$ and $B$ given a third
stimulus $X$ drawn from one of the two categories
\citep{schatz2016abx}. Applied to speech representations, ABX assumes
that the unit being evaluated corresponds approximately to a
segment-sized phonetic event: a single phoneme or a short acoustic
window aligned with it. This assumption holds reasonably well for
frame-level representations such as HuBERT \citep{hsu2021hubert} or wav2vec 2.0 \citep{baevski2020wav2vec}, which
operate at windows of 20--40 ms and can be aligned with
individual phonemes in controlled minimal-pair stimuli.

Mimi, however, operates at a fixed frame rate of 12.5 Hz, producing
one token every 80 ms. At this temporal granularity, a single token
routinely spans multiple phonemes: for example, the mean phoneme
duration in our TIMIT subset is 81.6 ms for SA1 and 73.9 ms for SA2,
with standard deviations exceeding 45 ms in both cases. As a result,
the token corresponding to a target phoneme in an ABX trial will
systematically also contain acoustic information from neighbouring
segments. The ABX score therefore reflects not only the phonetic
category of interest but also its phonetic context, making it
impossible to isolate the contribution of the target phoneme from
coarticulatory effects. This confound is not incidental: it is a
direct consequence of the 80 ms tokenization window, and it affects
all ABX evaluations of Mimi uniformly.

Posteriorgram-based probing avoids this confound by operating at the
level of statistical distributions rather than individual stimulus
comparisons. A phonetic posteriorgram assigns to each time frame a
probability vector over the full phoneme inventory, derived from a
separately trained acoustic model \citep{churchwell2024high}.
Rather than asking whether two tokens are discriminable in a pairwise
sense, we ask what phoneme distribution a given token is statistically
associated with across many speakers and contexts. This distributional
framing is robust to the 80 ms window because it does not require
precise phoneme boundary alignment: overlapping phoneme contributions
are averaged out across the many token--phoneme co-occurrences. The result is a principled measure of token--phoneme association that is not artificially inflated or deflated by the temporal granularity of the codec.

\section{Methodology}

\subsection{Data}

Our analysis relies on the TIMIT corpus \citep{garofolo1993darpa}, a widely used benchmark for phonetic research containing read English speech from 630 speakers across eight dialect regions of the United States. To enable controlled cross-speaker comparisons, we restrict our analysis to the two sentences read by all speakers (SA1 and SA2), which cover a broad range of English phonetic contexts, including stressed vowels, liquids, stops, and fricatives.

\begin{figure*}[htbp]
    \centering
    \begin{subfigure}[b]{0.98\columnwidth}
        \centering
        \includegraphics[width=\linewidth]{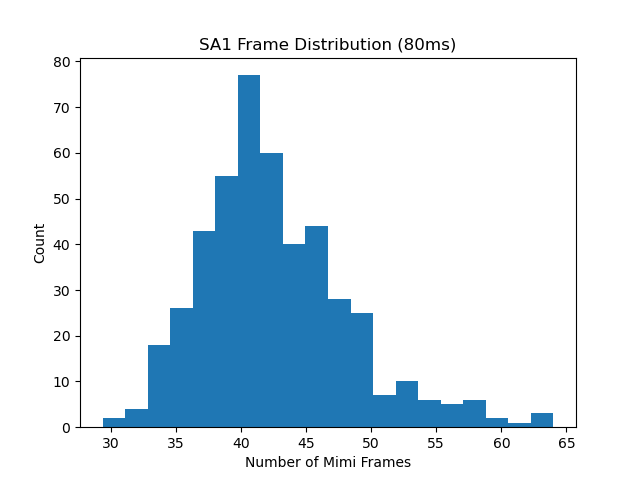}
        \caption{SA1}
        \label{fig:sa1_frames}
    \end{subfigure}
    \hfill
    \begin{subfigure}[b]{0.98\columnwidth}
        \centering
        \includegraphics[width=\linewidth]{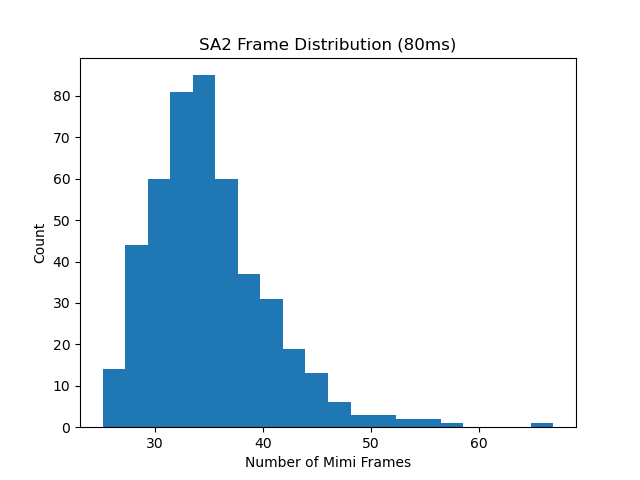}
        \caption{SA2}
        \label{fig:sa2_frames}
    \end{subfigure}
    \caption{Distribution of Mimi token frames (80\,ms each) in TIMIT
    SA1/SA2 across speakers. The modal counts (42 for SA1, 35 for SA2)
    are indicated by the tallest bar in each histogram.}
    \label{fig:frame_distributions}
\end{figure*}

Since Mimi operates at a fixed frame rate of 12.5 Hz (one token every 80 ms), utterance duration directly determines the number of tokens produced. We first examined the distribution of frame counts across all 630 speakers for each sentence (see Figure~\ref{fig:frame_distributions}). To ensure strict
positional comparability across speakers — a prerequisite for computing cross-speaker token distributions at aligned temporal positions — we retained only utterances whose token sequence length matched the modal frame count: 42 frames for SA1 and 35 frames for SA2. This yielded 58 speakers for SA1 and 55 speakers for SA2.

This filtering strategy reduces alignment ambiguity but introduces a
selection effect: speakers whose speech rate deviates from the modal duration are excluded. To assess the sensitivity of our results to this choice, we verified that the modal counts (42 and 35) correspond to the most frequent frame lengths by a substantial margin (see Figure~\ref{fig:frame_distributions}), and that the excluded speakers do not cluster systematically by dialect region or gender. We therefore treat the modal-length subsets as representative of the broader speaker population, while acknowledging that replication on larger sentence sets would strengthen this claim.

Phoneme boundaries were obtained using the Montreal Forced Aligner
\citep{mcauliffe2017montreal}, producing phone-level time intervals that were subsequently mapped onto Mimi token indices by maximum overlap assignment: each token was labelled with the phoneme occupying the largest proportion of its 80 ms window. The resulting dataset links each token occurrence to its speaker, utterance, temporal position, token identifier, and aligned phoneme label, providing the basis for all distributional analyses reported in Section \ref{sec:results}.

\subsection{Token Manipulation and Resynthesis Validation}

To complement the distributional analyses, we conducted a series of controlled interventions on Mimi's discrete representations prior to waveform reconstruction. Speech signals from the TIMIT corpus \citep{garofolo1993darpa} were encoded using Mimi, yielding eight parallel codebook streams. Modified token sequences were then decoded back into waveforms for analysis.

Two forms of intervention were considered. First, exploratory single-frame token substitutions were performed to qualitatively assess the local sensitivity of the reconstruction process to individual token changes. Second, a global replacement condition was applied in which all tokens from a selected codebook were replaced by a constant token value throughout the utterance. This intervention provides a stronger perturbation of the encoded representation and forms the basis of the quantitative evaluation reported in Section~4.5.

To assess the linguistic consequences of these manipulations, reconstructed waveforms were transcribed using a pretrained wav2vec~2.0 automatic speech recognition (ASR) system \citep{baevski2020wav2vec}. Word Error Rate (WER) was then computed between the reference transcripts and ASR outputs. Unlike perceptual inspection alone, WER provides an objective measure of the extent to which linguistic information is preserved following codebook manipulation \citep{ali2018word}.

\subsection{Forced Alignment}

To relate Mimi tokens to phonetic categories, we align discrete token sequences with phone-level transcriptions obtained through forced alignment. Phonetic boundaries are generated using the Montreal Forced Aligner (MFA), which produces time-aligned phoneme intervals for each utterance \citep{mcauliffe2017montreal}. Phoneme durations derived from MFA alignments exhibited substantial variability across speakers and contexts. For SA1, the mean phoneme duration was 81.6\,ms (median 80.0\,ms), while for SA2 it was 73.9\,ms (median 60.0\,ms). Although these values are comparable to Mimi's fixed token duration of 80\,ms, the large observed variability (standard deviations of 48.5\,ms and 47.6\,ms, respectively) implies that token and phoneme boundaries do not consistently coincide. We therefore employ an overlap-based token--phoneme assignment procedure rather than assuming a one-to-one correspondence between tokens and phonemes.

Because Mimi operates at a fixed frame rate of 12.5\,Hz, each token corresponds to an 80\,ms interval. For every token, we compute its temporal overlap with the MFA phoneme intervals and assign the phoneme occupying the largest proportion of the token duration. This procedure produces a token--phoneme alignment table linking each token occurrence to an aligned phonetic category. The resulting dataset contains token identity, token position, speaker metadata, utterance information, and aligned phoneme labels.

\section{Results}
\label{sec:results}

\subsection{Duration and Token Distribution Across Speakers}

To assess cross-speaker stability of the first codebook, we compute
the entropy of token assignments at each temporal position across
speakers within the modal-length subsets (58 speakers for SA1,
55 speakers for SA2):
\[
H(X) = -\sum_{i} p(x_i) \log_2 p(x_i)
\]
where $p(x_i)$ denotes the probability of observing token $x_i$ at a
given frame position. Mean entropy is 3.79 bits for SA1 and 4.30 bits
for SA2, substantially below the theoretical maximum of 11 bits for a
2,048-token codebook. This suggests that token assignments are
structured across speakers at corresponding temporal positions, rather
than uniformly distributed.

We note, however, that entropy alone is insufficient to determine
whether this structure reflects genuine phonetic organization, since a
codec optimised for compression is expected by design to distribute
token usage broadly across the codebook. A direct measure of
token--phoneme correspondence, independent of this compression
objective, is provided by the PNMI analysis reported in
Section~\ref{sec:pnmi}, which confirms that the observed structure
is substantially above what would be expected under random token
assignment.

\subsection{Token--Phoneme Alignment under Controlled Temporal Conditions}

Using Montreal Forced Aligner phone boundaries aligned to the first
codebook tokens (modal-length subsets: 58 speakers for SA1, 55 for
SA2), we computed the conditional distribution
$P(\text{phoneme} \mid \text{token})$, token entropy, and dominant
phoneme probability. The results show very low entropy values: 0.31
bits for SA1 and 0.33 bits for SA2, substantially below the
theoretical maximum of 11 bits for a 2,048-token codebook.  
The mean dominant phoneme probability is approximately 0.89 in both cases, indicating that most tokens are strongly associated with a single phoneme. More than 170 tokens exhibit a dominant probability above 0.6, and nearly 200 tokens have entropy below 2 bits.

These findings suggest that the first codebook tokens exhibit highly
concentrated phoneme distributions across speakers. We note that low
entropy of $P(\text{phoneme} \mid \text{token})$ is not trivially
expected from compression design: a codec optimised to spread usage
uniformly across 2,048 tokens would produce high token entropy, not
low phoneme entropy conditional on each token. The PNMI baseline
reported in Section~\ref{sec:pnmi} further confirms that this
concentration exceeds what would be expected under random token
assignment ($3.8\times$ and $2.9\times$ above chance for SA1 and SA2
respectively). Although the association is not strictly one-to-one,
the low entropy values indicate that most tokens are reliably
dominated by a single phonetic category.

\subsection{Phoneme--Token Alignment}
\label{phoneme_token}

To evaluate whether each phoneme is associated with a dedicated set
of tokens, we computed the conditional distribution
$P(\text{token} \mid \text{phoneme})$ for the controlled SA1 and SA2
subsets. In contrast to the low entropy observed in the reverse
direction, phoneme entropy is substantially higher: 2.58 bits for SA1
and 2.63 bits for SA2. The mean dominant token probability is
approximately 0.39 in both cases, and only a small number of phonemes
exhibit a dominant token probability above 0.6.

\begin{figure}[t]
    \centering
    \begin{subfigure}{0.45\textwidth}
        \centering
        \includegraphics[width=\linewidth]{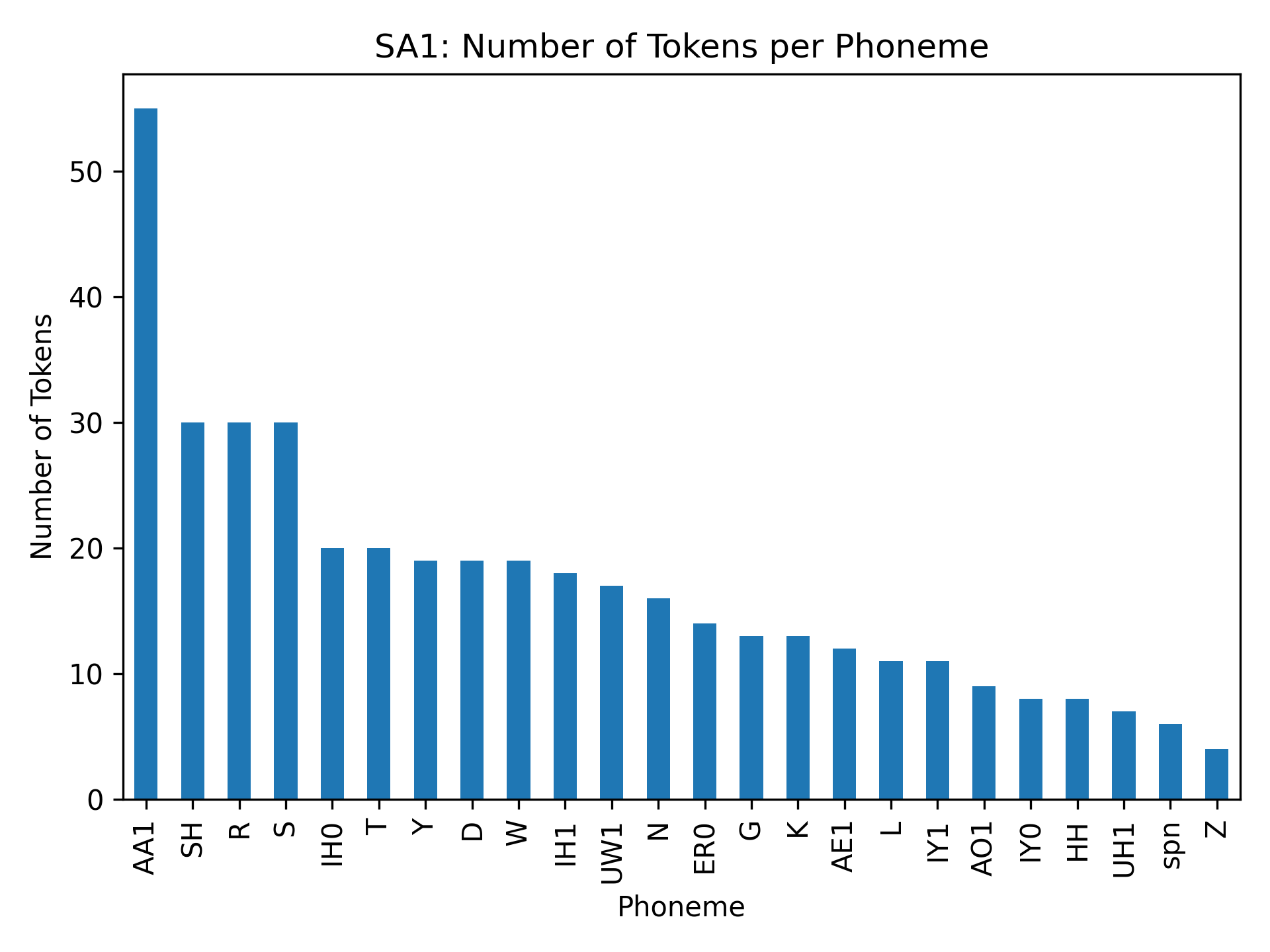}
        \caption{Token variability (SA1)}
        \label{fig:sa1_token_variability}
    \end{subfigure}
    \hfill
    \begin{subfigure}{0.45\textwidth}
        \centering
        \includegraphics[width=\linewidth]{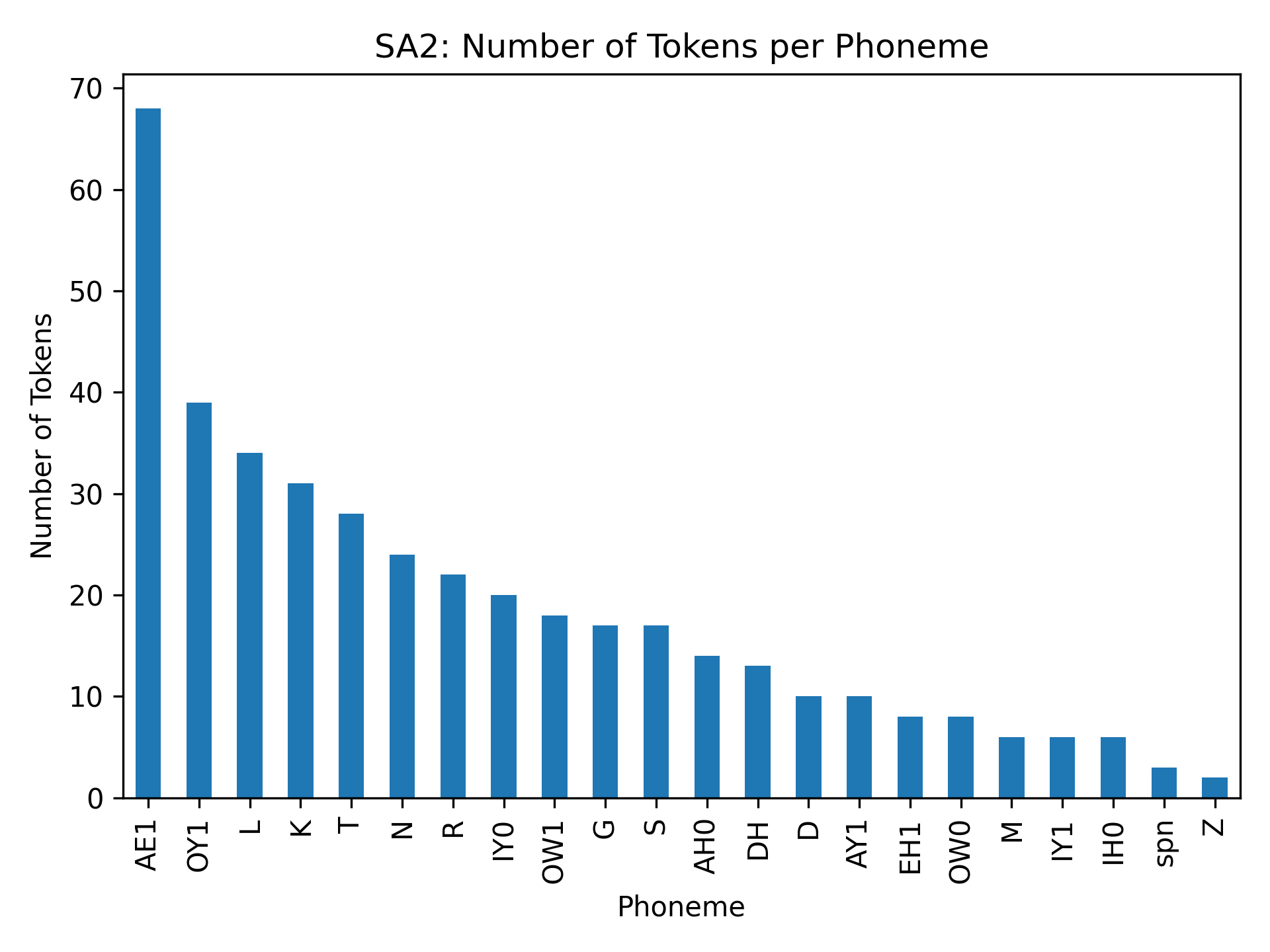}
        \caption{Token variability (SA2)}
        \label{fig:sa1_pnmi_scatter}
    \end{subfigure}
    \caption{Token variability in Mimi's first codebook across TIMIT SA1 and SA2}
    \label{fig:sa1_dual}
\end{figure}

These results indicate a strong asymmetry in the token--phoneme
mapping (see Figure~\ref{fig:sa1_token_variability}). While individual
tokens are highly consistent with specific phonemes, individual
phonemes are realized by multiple competing tokens. Given the ratio of
2,048 tokens to approximately 25 observed phonemes in our data, some
degree of $P(\text{token} \mid \text{phoneme})$ entropy is expected by
construction: each phoneme could in principle be realized by up to
$2048/25 \approx 82$ tokens under uniform assignment. The observed
entropy values of 2.58 and 2.63 bits correspond to effective
vocabulary sizes of approximately $2^{2.58} \approx 6$ and
$2^{2.63} \approx 6$ competing tokens per phoneme on average,
substantially fewer than the 82 that would be expected under uniform
distribution, indicating that the competition is concentrated rather
than random. This pattern is consistent with an allophonic encoding
in which contextual variants of a phoneme are captured by distinct
but related tokens, rather than a minimal phoneme-level
discretization.

\subsection{Cross-sentence stability of token families}
\label{sec:jsd}

To determine whether the observed cross-sentence variability could be explained by finite sampling, we conducted a phoneme-conditioned permutation test. For each phoneme occurring in both SA1 and SA2, we computed the token distributions conditioned on that phoneme in each sentence and quantified their similarity using Jensen--Shannon divergence (JSD).

A potential criticism of this comparison is that differences between SA1 and SA2 may simply arise from limited sample sizes. To address this possibility, we constructed an empirical null distribution. For each phoneme, all token observations from the two sentences were pooled and randomly repartitioned into groups matching the original sample sizes. JSD was then recomputed on the permuted samples. This procedure was repeated 1,000 times to estimate the divergence expected under the null hypothesis that both sentences were generated from the same underlying token distribution.

The observed divergences substantially exceeded the permutation baseline. The mean observed JSD across the 15 phonemes shared by both sentences was 0.61, whereas the mean permutation JSD was only 0.14. Fourteen of the fifteen phonemes exhibited significant divergence under the empirical null hypothesis ($p<0.05$), and all remained significant after Benjamini--Hochberg correction for multiple comparisons.

Several phonemes, including /G/, /IY/, /IH/, /S/, and /R/, exhibited JSD values approaching the theoretical maximum of $\ln 2$, indicating an almost complete redistribution of token families across the two sentences. These findings demonstrate that the instability of token families cannot be explained solely by finite-sampling variability. Instead, the first Mimi codebook appears to encode information extending beyond broad phonemic identity, preserving fine-grained aspects of speech realization that vary across sentence contexts.

\begin{table}[t]
\centering
\caption{Permutation test results.}
\label{tab:jsd_perm}
\small
\begin{tabular}{lcccc}
\hline
Phon. & Obs. & Null & 95\% & $p$ \\
\hline
G   & 0.693 & 0.166 & 0.214 & .001 \\
IY0 & 0.693 & 0.089 & 0.130 & .001 \\
IH0 & 0.693 & 0.213 & 0.282 & .001 \\
S   & 0.693 & 0.108 & 0.137 & .001 \\
R   & 0.689 & 0.057 & 0.077 & .001 \\
K   & 0.668 & 0.114 & 0.150 & .001 \\
AE1 & 0.668 & 0.122 & 0.153 & .001 \\
T   & 0.626 & 0.191 & 0.250 & .001 \\
OW0 & 0.284 & 0.214 & 0.286 & .179 \\
Z   & 0.187 & 0.101 & 0.149 & .034 \\
\hline
\end{tabular}
\end{table}

\subsection{PNMI Analysis}
\label{sec:pnmi} 

Following \cite{zhang2024speechtokenizer}, we compute PNMI as
$\text{PNMI} = I(\text{token}; \text{phoneme}) / H(\text{phoneme})$,
where higher values indicate stronger token--phoneme correspondence.
Restricting to modal-length subsets (42 frames for SA1, $n=58$;
35 frames for SA2, $n=55$), Mimi's first codebook achieves PNMI~$= 0.84$ for both sentences. To verify this is not a compression artefact, we computed a random baseline by shuffling token assignments
(100 iterations): $0.22 \pm 0.004$ for SA1 and $0.29 \pm 0.004$ for SA2, placing the observed values $3.8\times$ and $2.9\times$ above chance respectively. Table~\ref{tab:pnmi} compares these results against values reported by \cite{zhang2024speechtokenizer} on the full TIMIT training set. Mimi CB0 outperforms both EnCodec and SpeechTokenizer RVQ-1 despite lacking an explicit phoneme-level training objective, consistent with its WavLM distillation \citep{defossez2024moshi}. Direct comparison is indicative, given differences in dataset size.

\subsection{Phonological Information Across the Mimi Codebooks}

The previous analyses focused exclusively on the first Mimi codebook (CB0), following the architectural assumption that the first codebook captures the highest-level linguistic information. To evaluate whether phonological structure is restricted to CB0 or distributed across the residual vector quantization (RVQ) hierarchy, we repeated the PNMI analysis for all eight codebooks.

\begin{table}[ht]
\centering
\small
\begin{tabular}{l l c c}
\hline
\textbf{Model} & \textbf{Layer} & \textbf{PNMI} & \textbf{Baseline} \\
\hline
HuBERT         & KM500 & 0.43 & ---              \\
EnCodec         & RVQ-1 & 0.28 & ---              \\
SpeechTokenizer & RVQ-1 & 0.71 & ---              \\
\hline
Mimi (SA1)                & CB0   & \textbf{0.84} & $0.22 \pm 0.004$ \\
Mimi (SA2)                & CB0   & \textbf{0.84} & $0.29 \pm 0.004$ \\
\hline
\end{tabular}
\caption{PNMI of discrete speech representations.
Values for HuBERT, EnCodec, and SpeechTokenizer are
from \cite{zhang2024speechtokenizer}, Table 8, computed
on the full TIMIT training set. Mimi values are computed
on the modal-length SA1/SA2 subsets; Mimi baseline:
token shuffling, 100 iterations.}
\label{tab:pnmi}
\end{table}

Table~\ref{tab:codebook_pnmi} reports the PNMI values obtained for each codebook. As expected, the first codebook exhibits the strongest correspondence with phoneme categories, reaching PNMI values of approximately 0.84 for both SA1 and SA2. However, the remaining codebooks also display substantial phoneme-related information, with PNMI values ranging between approximately 0.55 and 0.67.

These results indicate that phonological information is not exclusively concentrated in the first codebook. Instead, segmental structure remains accessible throughout the RVQ hierarchy, although with a reduced degree of organization relative to CB0. This finding suggests that later codebooks do not merely encode waveform reconstruction residuals, but preserve information that remains predictive of phonetic categories.

The result is consistent with the hypothesis that speech representations in Mimi are distributed across multiple codebooks rather than being localized within a single discrete representation layer. While CB0 provides the strongest phonological signal, subsequent codebooks continue to encode information relevant to phonetic identity.

\begin{table}[t]
\centering
\caption{
Phoneme Normalized Mutual Information (PNMI) measured independently for each Mimi codebook.
CB0 exhibits the strongest phonological correspondence, while all remaining codebooks retain
substantial phoneme-predictive information.
}
\label{tab:codebook_pnmi}
\begin{tabular}{lcc}
\hline
Codebook & SA1 PNMI & SA2 PNMI \\
\hline
CB0 & 0.8407 & 0.8394 \\
CB1 & 0.6203 & 0.5566 \\
CB2 & 0.6415 & 0.6613 \\
CB3 & 0.6260 & 0.6583 \\
CB4 & 0.6218 & 0.6648 \\
CB5 & 0.6296 & 0.6541 \\
CB6 & 0.6365 & 0.6657 \\
CB7 & 0.6288 & 0.6711 \\
\hline
\end{tabular}
\end{table}

\subsection{Controlled Interventions}

We evaluate the perceptual and linguistic impact of token manipulations on utterances from the TIMIT corpus encoded with Mimi.
To isolate the contribution of codebook\_0, we define four
experimental conditions: (i)~the unmodified original waveform,
(ii)~a Mimi encode--decode reconstruction without any token modification,(iii)~replacement of a single frame in codebook\_0 with a
constant token identifier, and (iv)~global replacement of all tokens in codebook\_0 with the same constant value.
Conditions (iii) and (iv) are decoded by Mimi after modification,
yielding resynthesized waveforms that reflect each intervention.
All waveforms were transcribed using a pretrained wav2vec~2.0 ASR model \citep{baevski2020wav2vec}, and Word Error Rate (WER) was computed against the TIMIT reference transcriptions as a measure of phonetic and lexical divergence. Results are reported on SI and SX sentences ($n = 8$) from a single speaker (FCJF0); SA sentences are excluded due to their known difficulty for automatic speech recognition systems \citep{garofolo1993darpa}.

\begin{table}[ht]
\centering
\caption{WER under four token manipulation conditions
         (SI/SX, speaker FCJF0, $n=8$).}
\label{tab:wer_conditions}
\small
\begin{tabular}{lc}
\toprule
\textbf{Condition} & \textbf{WER} \\
\midrule
Original audio                           & 0.194 \\
Mimi reconstruction                      & 0.493 \\
CB\textsubscript{0} single token         & 0.493 \\
CB\textsubscript{0} all tokens           & 0.701 \\
\bottomrule
\end{tabular}
\end{table}

Table~\ref{tab:wer_conditions} reveals three distinct patterns.
First, Mimi reconstruction alone raises WER from 0.194 to 0.493,
reflecting a domain mismatch between TIMIT read speech and Mimi's
training distribution. This reconstruction baseline is essential
for interpreting subsequent manipulations: any degradation
attributable to the intervention must be evaluated relative to
this baseline, not to the original signal.

Second, replacing a single frame in codebook\_0
produces no additional WER increase beyond reconstruction
(WER = 0.493), indicating that isolated token perturbations are
absorbed by the remaining seven codebooks and do not propagate
to intelligibility loss. This null result suggests that the
contribution of codebook\_0 is distributed across
frames rather than concentrated in individual tokens.

Third, global replacement of all tokens in codebook\_0 with a constant value causes a further
increase to WER = 0.701, a degradation of $+0.208$ relative to
the reconstruction baseline. While the resulting waveforms remain
perceptually speech-like, the ASR degradation demonstrates that
codebook\_0 carries linguistically relevant
structure that, when systematically disrupted, produces measurable
phonetic and lexical divergence. This result is consistent with
the hypothesis that the first codebook encodes phonemically
relevant information, complementing the distributional findings
reported in Sections~4.2 and~4.3.

\section{Discussion}
\subsection{Allophonic Realizations}

To determine whether token competition reflects classical phonological
allophony, we examined the empirical distribution of surface phoneme
labels per canonical category in the SA1 and SA2 subsets. Stress
variants for vowels were collapsed (e.g., IH0/IH1) in order to focus
on potential segmental alternations rather than prosodic differences.

\begin{table}[h]
\centering
\small
\begin{tabular}{lccc}
\hline
Phoneme & SA1 Entropy & SA2 Entropy & Distinct  \\
        & (bits)      & (bits)      & Tokens             \\
\hline
L   & 1.70 & 3.93 & 3--18 \\
K   & 1.28 & 3.36 & 4--17 \\
T   & 2.12 & 3.60 & 6--17 \\
R   & 2.99 & 3.00 & 9--21 \\
AE1 & 1.78 & 4.68 & 5--35 \\
\hline
\end{tabular}
\caption{Representative phoneme entropy values for the conditional
distribution $P(\text{token}\mid\text{phoneme})$ in SA1 and SA2.
Low entropy indicates that a phoneme is realized by a small set of
tokens; higher entropy reflects greater token variability.}
\label{tab:phoneme_entropy}
\end{table}

Phoneme annotations follow the ARPAbet conventions used in the CMU
Pronouncing Dictionary and were obtained through forced alignment with
the Montreal Forced Aligner. Under these conventions, most phonemes
are represented categorically, and the transcription does not
explicitly encode many contextual allophonic processes such as
flapping or aspiration. As a result, aside from stress variation, the
aligned phoneme labels show little surface alternation across speakers.

Despite this categorical phonemic representation, the Mimi tokens
associated with a given phoneme exhibit substantial variability across
speakers (see Table~\ref{tab:phoneme_entropy}). As established in
Section~\ref{phoneme_token}, the mean effective competitor set
per phoneme is approximately 6 tokens ($2^{2.58}$ for SA1), far below
the 82 tokens expected under uniform random assignment. This indicates
that token competition is concentrated and structured, even if it
exceeds a strict one-to-one phoneme--token mapping.

Notably, entropy values differ considerably between SA1 and SA2 for
the same phoneme, for instance, /L/ rises from 1.70 bits in SA1
to 3.93 bits in SA2, and /AE1/ from 1.78 to 4.68 bits. This
cross-sentence variability is consistent with the permutation test
results reported in Section~\ref{sec:jsd}, which demonstrated that
token family distributions for the same phoneme differ significantly
across sentence contexts (mean JSD = 0.61, p $<$ 0.05 for 14 of 15
shared phonemes). Rather than reflecting random sampling noise, this
instability suggests that Mimi tokens are sensitive to phonetic
context beyond the phoneme label itself.

A plausible interpretation is that Mimi tokens capture fine-grained
acoustic differences driven by coarticulation and reduction effects,
particularly in intervocalic environments. Since these contextual
variations are not explicitly encoded in CMU ARPAbet labels, they
surface instead as token-level variability. This is consistent with
an allophonic encoding: the first codebook does not maintain a stable inventory of phoneme-sized units, but instead reflects the surface acoustic realization of each segment in its specific phonetic context.



\section{Conclusion}
By focusing the analysis on the two utterances pronounced by all the speakers of the TIMIT corpus, we were able to show correlations between Mimi's first codebook tokens with allophonic rather than phonemic representations. 

In this respect, when probing Mimi for phon* representations at the phonetic-phonology interface,  we tend to conclude that the Mimi codec as a speech tokenizer seems to yield phonetic categories in its first codebook tokens.
Nevertheless, the wider 80 ms temporal window of Mimi's tokens, compared to codecs operating at higher frame rates such as SpeechTokenizer, raises the possibility that they also capture speaker-related acoustic detail alongside phonetic content.

\section*{Limitations}
An obvious limitation is we focused on the two utterances pronounced by all the speakers of the TIMIT corpus. The distribution of tokens as allophonic representation should be tested with more diverse varieties of English such as the ones present in the Edinburgh corpus of English \citep{sanabria2023edinburgh}.

\section*{Acknowledgment}
We thank Guillaume Wisniewski and anonymous reviewers for comments on an earlier version of this paper. We resorted to Claude for part of the code or for reformulations. Authors manually checked the LLM outputs. Data was processed using the platform PAPTAN, funded by a 2021 research equipment grant from the Scientific Platforms and Equipment Committee (PAPTAN project), under ANR Grant Number ANR-18-IDEX-0001 (Financement IdEx Université de Paris).

\bibliography{custom}

\appendix



\end{document}